\documentclass[smallcondensed]{svjour3}
\usepackage{graphicx} 
\usepackage{natbib}
\usepackage{subfigure}

\usepackage{booktabs}
\usepackage{subfigure}
\usepackage{amsmath}
\usepackage{geometry,graphicx}
\usepackage{bbm}
\usepackage{enumerate}
\usepackage[table]{xcolor}
\definecolor{lightblue}{RGB}{230,247,254}
\usepackage{graphicx}
\usepackage{enumitem}
\usepackage{soul,color,url}
\usepackage{multirow,bm,subfigure,tikz}
\newtheorem{Proposition}{Proposition}

\begin{document} 
      \pgfmathdeclarefunction{gauss}{3}{%
	      \pgfmathparse{1/(#3*sqrt(2*pi))*exp(-((#1-#2)^2)/(2*#3^2))}%
      }
      \title{Statistical comparison of classifiers through Bayesian hierarchical modelling}
      \author{Giorgio Corani \and Alessio Benavoli \and Janez Dem{\v{s}}ar \and Francesca Mangili \and Marco Zaffalon}
      \institute{G. Corani,  A. Benavoli F. Mangili and M. Zaffalon  are with \at
      	Istituto Dalle Molle di studi sull'Intelligenza Artificiale (IDSIA),
      	Manno, Switzerland\\
      	J. Dem{\v{s}}ar is with  \\ Faculty of Computer and Information Science, University of Ljubljana,
        		  Slovenia\\
      	\email{\\giorgio\{alessio,francesca,zaffalon\}@idsia.ch\\janez.demsar@fri.uni-lj.si	
      	}           
      }
      \date{Received: date / Accepted: date}
      \maketitle

\begin{abstract} 
Usually one compares
the accuracy of two competing classifiers via null hypothesis significance tests (nhst).
Yet the nhst tests
suffer from important shortcomings, which
can be overcome by switching to Bayesian hypothesis testing.
We propose a Bayesian hierarchical model which
jointly analyzes the cross-validation results obtained by two classifiers on multiple data sets.
It returns the posterior probability 
of the accuracies of the two classifiers being practically
equivalent or significantly different.
A further strength of the hierarchical model is that, by jointly analyzing the results obtained on all data sets,
it reduces the estimation error compared to the usual approach of averaging the cross-validation results obtained on a given data set. 
\end{abstract}

\section{Introduction}

The statistical comparison of competing algorithms is fundamental in machine learning; it is typically carried through hypothesis testing. As an example in this paper we assume that one is interested in comparing the accuracy of two competing classifiers. However our discussion readily applies to any other measure of performance.

Assume that two classifiers have been assessed via cross-validation on a single data set. The recommended approach for comparing them is the correlated t-test \citep{nadeau2003inference}. If instead one aims at comparing two classifiers on multiple data sets the recommended test is the signed-rank test \citep{demvsar2006statistical}. Both tests are based on the frequentist framework of the null-hypothesis significance tests (nhst), which has severe drawbacks.

First, the nhst computes the probability of getting the observed (or a larger) difference in the data if the null hypothesis was true. It does \textit{not} compute
the probability of interest, which is the probability
of one classifier being more accurate than another given the observed results. 

Second, the claimed statistical significances do not necessarily imply practical significance.
It has been pointed out  \citep{hand2006classifier} that the apparent superiority of 
a classifier in a simulation may be swamped by other sources of uncertainty when the classifier is adopted in practice.
This especially applies when the new classifier provides only a thin advantage over the previous one.
Yet the signed-rank test easily rejects the null hypothesis 
even when dealing with two classifiers separated by a thin difference,
if the classifiers have been compared on 
a large collection of data sets. 
Null hypotheses can virtually always be rejected by increasing the sample size \citep{wasserstein2016asa}, since
the p-value depends  both on the actual difference between the two classifiers and number of collected observations.

Another issue is that when the nhst does not reject the null, it provides no 
evidence in favor of the null hypothesis \citep[Chap.~11]{kruschke2015doing}:
it simply states that  there is not enough evidence for rejecting  the null hypothesis.
This prevents nhst tests from recognizing equivalent classifiers.

These issues can be overcome by  switching
to Bayesian hypothesis testing, as exhaustively discussed by \cite[Sec. 11]{kruschke2015doing}.
Recent applications of 
Bayesian hypothesis testing in machine learning have been proposed in
\citep{lacoste2012bayesian, coraniML2015};
yet there is currently no Bayesian approach able to jointly analyze the 
cross-validation results obtained 
on multiple data sets.

Let us introduce some notation.
We have a collection of $q$ data sets; the actual mean difference of accuracy on the
i-th data set is $\delta_i$.
We can think of $\delta_i$ as the average difference of accuracy  that we would obtain by repeating many times the procedure
of sampling from the actual distribution as many instances as there are in the i-data set available to us, train the two classifiers
and measure their difference of accuracy on a large test set.
We cannot know the actual value of $\delta_i$; thus we estimate
it through cross-validation. The usual estimate of $\delta_i$ is the mean of the cross-validation results
obtained on the i-th data set; this is also the maximum likelihood estimator (MLE) of 
$\delta_i$.

We propose the first model that represents both
the distribution $p(\delta_i)$ across the different data sets and
the distribution of the cross-validation results on the i-th data set given
$\delta_i$, modelling both the variance and the correlation of the cross-validation results on each data set.

In order to detect equivalent classifiers, we introduce
a region of practical equivalence (rope), as in \cite{kruschke2013bayesian}.   
In particular we consider two classifiers to be 
practically equivalent if their difference of accuracy belongs to the interval
$(-0.01, 0.01)$.
This constitutes a sensible default, even if there are no uniquely correct rope limits. 
We analyze how much of the posterior probability mass lies within
the rope,
at its left and at its right  in order to detect classifiers which
are practically equivalent or instead
significantly different.
With this, we can predict the probability that the $\delta_i$ new data set will lie in the rope or to the left or right of it.

Compared to the signed-rank, the hierarchical test is more conservative as it rejects less easily the null hypothesis.
The signed-rank assumes a point null hypothesis, while the 
null hypothesis of the hierarchical model contains all the values of the rope.
Such a null hypothesis is more realistic and thus more difficult to reject.

Consequently, the hierarchical test tend to be less powerful and more likely to retain the null-hypothesis when the two classifiers are in fact different. Yet this issue is mitigated by the fact that the posterior probabilities can be meaningfully interpreted even when they do not exceed the 95\% threshold: the result of the test is not a binary decision but a probability.

A further strength of the hierarchical model is that it reduces the estimation
error of the $\delta_i$'s compared to the traditional MLE approach.
The hierarchical model 
estimates the distribution from which the $\delta_i$'s are sampled.
Then it jointly estimates the values of the $\delta_i$'s.
By doing so it applies \textit{shrinkage},
namely the estimates $\hat{\delta}_i$'
are closer to each other than in the MLE case.
It is known that the shrinkage estimator yields lower mean squared error than MLE in the i.i.d. case \citep[Sec 6.3.3.2]{murphy2012machine}.
We prove theoretically that the shrinkage estimator yields lower mean squared error than MLE 
also in the correlated case, such as that of cross-validation;
moreover, we verify this statement in a variety of different experiments.

\section{Existing approaches}\label{sec:infer-single-dset}

We have a collection of $q$ data sets; the actual mean difference of accuracy on the
i-th data set is $\delta_i$.
We can think of $\delta_i$ as the average difference of accuracy  that we would obtain by repeating many times the procedure
of sampling from the actual distribution as many instances as there are in the actually available data set, train the two classifiers and measure the difference of accuracy on a large test set.

In reality we cannot know $\delta_i$.
We can however estimate it through cross-validation.
Assume that we have performed $m$ runs of $k$-fold cross-validation on each data set, providing both classifiers with paired folds. 
The \textit{differences of accuracy} on each fold of cross-validation are $\bm{x_i}=\{x_{i1},x_{i2},\dots,x_{in}\}$, where $n=mk$.
The  mean and the standard deviation of the results on the i-th data set are
$\bar{x}_i$ and $s_i$.
The mean of the cross-validation results
is also the 
maximum likelihood estimator (MLE) of 
$\delta_i$. 

The values $\bm{x_i}=\{x_{i1},x_{i2},\dots,x_{in}\}$ are not independently sampled; 
they are  instead sampled with correlation $\rho$ because of the overlapping training sets built during cross-validation. 
\cite{nadeau2003inference} prove that there is no unbiased estimator of the correlation and they approximate it
as $\rho=\frac{1}{k}$, where $k$ is the number of folds.
They devise the correlated $t$-test, whose statistic is: 
\begin{equation}
t= \overline{x}_i/{\sqrt{\hat{s}^2_i(\frac{1}{n}+\frac{\rho}{1-\rho})}}. 
\label{eq:ttest-nadeau}
\end{equation}
The denominator of the statistic
is the standard error,
which is informative about 
the accuracy
of $\bar{x}_i$  as an estimator of  $\delta_i$.
The standard error of the correlated t-test accounts for the correlation of cross-validation results.
The statistic of Eqn.(\ref{eq:ttest-nadeau}) follows a t distribution with n-1 degrees of freedom.
When the statistic exceeds the critical value, the test claims that $\delta_i$ is significantly different from zero.
This is the standard approach for comparing two classifiers on a single data set.

The signed-rank test is instead the recommended method \citep{demvsar2006statistical} to compare two classifiers on a collection of $q$ different data sets.
It is usually applied after having 
performed cross-validation on each data set.
The test analyzes the  mean differences 
measured on each data set ($\bar{x}_1,\bar{x}_2,\ldots,\bar{x}_q$) assuming them to be i.i.d..
This is a simplistic assumption: the $\bar{x}_i$'s
are not i.i.d. since they are characterized by different uncertainty;
indeed their standard errors are typically different.

 	      The test statistic is:
 	      \begin{equation*} \label{eq:wsignstat}
 	      \begin{array}{rcl}
 	      T^+=&\sum\limits_{\{i:~\bar{x}_i\geq 0\}} r_i (|\bar{x}_i|)&= \sum\limits_{1\leq i \leq j \leq n} T^+_{ij},\vspace{2mm} \\
 	      \end{array}
 	      \end{equation*} 
 	      where $r_i (|\bar{x}_i|)$ is the rank of $|\bar{x}_i|$ and
 	      $$
 	      T^+_{ij}=\left\{\begin{array}{ll}
 	      1 & \textit{if } \bar{x}_i \geq \bar{x}_j,\\
 	      0 & \textit{otherwise. } \\
 	      \end{array}\right.
 	      $$
 	      For a large enough number of samples (e.g., $q$\textgreater 10), 
 	      the statistic  under the null hypothesis is  normally distributed.
 	      When the test rejects the null hypothesis, it claims that the median of the population of the $\delta_i$'s is different from zero.
 	  
 	  The two tests discussed so far are null-hypothesis significance test (nhst) and as such they suffer from the drawbacks 	  discussed in the introduction.

 	  Let us now consider the Bayesian approaches.
\cite{kruschke2013bayesian} presents a Bayesian t-test coupled with the rope for i.i.d. observations.
Because of the i.i.d. assumption, it
\textit{cannot} be applied to analyze the cross-validation results, which are correlated.
The Bayesian correlated t-test  by \cite{coraniML2015} 
can be used to this end.
It computes the posterior distribution of $\delta_i$ 
  on a \textit{single} data set, assuming the cross-validation  observations  
to be sampled from a multivariate normal distribution whose components 
have the same mean $\delta_i$, the same standard deviation $\sigma_i$ and  
are equally cross-correlated with correlation $\rho=\frac{1}{k}$.  
Thus the test borrows the correlation heuristic by \cite{nadeau2003inference}.

As for the analysis on multiple data sets,
\citet{lacoste2012bayesian} compares two classifiers  on multiple data sets by
modelling each data set as an independent Bernoulli trial. 
The two possible outcomes of the Bernoulli trial the first classifier being more accurate than the second or vice versa.
\cite{coraniML2015} estimate such probabilities using the Bayesian correlated t-test, applied independently on each data set.
It is then possible computing the probability of the first classifier 
being more accurate than the second classifier on more than half of the $q$ data sets.
A shortcoming of this approach is that 
its conclusions apply only to the $q$ available data sets without 
generalizing to the 
population of data sets from which they have been sampled.

\section{The hierarchical model}\label{sec:hier}
We propose a Bayesian hierarchical model for comparing two classifiers.
Its core assumptions are:
\begin{align}
&	\delta_1 ... \delta_q \sim t (\delta_0, \sigma_0,\nu ), \label{eq:delta_i} \\
& \sigma_1 ... \sigma_q \sim \mathrm{unif} (0,\bar{\sigma}), \label{eq:sigma_i} \\ 
&	\mathbf{x}_{i} \sim  MVN(\mathbf{1} \delta_i,\mathbf{\Sigma_i}).  \label{eq:mvn} 
\end{align}

The i-th data set is characterized by the mean 
difference of accuracy between classifiers 
$\delta_i$ and the standard deviation $\sigma_i$.
Thus we model each data sets as having its own estimation uncertainty.
Notice that instead when applying the signed-rank test one has to assume that
the $\bar{x}_i$'s are i.i.d. 

The $\delta_i$'s are assumed to be drawn from a Student distribution
with mean  $\delta_0$, scale factor $\sigma_0$ and degrees of freedom $\nu$.
We choose the Student distribution because it is more flexible than the Gaussian,
thanks to the additional parameter $\nu$. 
When $\nu$ is small, the  Student distribution has heavy tails; when  $\nu$ is above 30, the  Student distribution is practically a Gaussian.
A Student distribution with low degrees of freedom is able to robustly
model outlier data points, namely some
$\delta_i$'s which are far away from the others. 
Indeed the Student distribution is 
typically used for robust Bayesian estimation \citep{kruschke2013bayesian}. 

We assume $\sigma_i$ to be drawn from a uniform distribution over the interval $(0,\bar{\sigma})$. 
This kind of prior is recommended for the standard deviation by \cite{gelman2006prior}, as it yields inferences which are insensitive on the value of $\bar{\sigma}$ if $\bar{\sigma}$ is large enough. 
We  adopt 
$\bar{\sigma}= 1000 \bar{s}$  where $\bar{s}$ is the mean standard deviation observed on the different data sets,
$\bar{s}=\sum_i^q s_i/q$.
As we show in Sec.~\ref{sec:sens-sigma}, we obtain the same 
posterior distribution if we change the upper bound to $100\bar{s}$.

Equation (\ref{eq:mvn}) models the fact that the cross-validation measures $\bm{x_i}=\{x_{i1},x_{i2},\dots,x_{in}\}$  of the i-th data set are  generated 
from a multivariate normal whose components are equally cross-correlated with correlation $\rho$, have same mean ($ \delta_i$) and same standard deviation ($\sigma_i$).
These are standard assumptions for the cross-validation measures \citep{nadeau2003inference, coraniML2015}.
The normality assumption is sound since the average accuracy over the instances of the test set 
tends to be normally distributed by the central limit theorem.

We complete the model with the prior on the parameters 
$\delta_0$, $\sigma_0$ and $\nu$
of the high-level distribution.
We assume  $\delta_0$ 
to be uniformly distributed
within 1 and -1. This choice works for 
all the measures  bounded within $\pm$1, such as 
accuracy, AUC, precision and recall. Other type of indicators might require different bounds.

For the standard deviation
$\sigma_0$ we adopt the prior  $unif(0,\bar{s_0})$, with
$\bar{s_0}=1000 s_{\bar{x}}$, where $s_{\bar{x}}$ is the  standard deviation of the 
$\bar{x}_i$'s. Also in this case we checked that the posterior distribution of 
$\sigma_0$ is unchanged if we adopt    
$\bar{s_0}=1000 s_{\bar{x}}$ or $\bar{s_0}=100 s_{\bar{x}}$   as upper bound.

The treatment of $p(\nu)$ is slightly more challenging.   
\cite{kruschke2013bayesian} 
proposes
$p(\nu)$ to be a shifted exponential, which we re-parameterize as a 
Gamma($\alpha$,$\beta$) with
$\alpha$=1, $\beta$= 0.0345.
\cite{juarez2010model} proposes instead 
$p(\nu) = \mathrm{Gamma}(2,0.1)$. 
The main characteristics of those distributions
are given in Tab.~\ref{tab:prior-nu}.

The inferred model shows some sensitivity on the choice of $p(\nu)$.
Yet both parameterizations are sensible and we have no strong reason to prefer one over another.
We model this uncertainty by representing  the coefficients $\alpha$ and $\beta$ as two random variables with their own prior distribution (hierarchical approach).
In particular we assume
$p(\nu)=\mathrm{Gamma}(\alpha,\beta)$, with 
$\alpha \sim \mathrm{unif} (\underline{\alpha},\bar{\alpha})$ and $\beta \sim \mathrm{unif} (\underline{\beta}, \bar{\beta})$, setting 
$\underline{\alpha}$=0.5, $\bar{\alpha}$=5, $\underline{\beta}$=0.05, $\bar{\beta}$=0.15.
The mean and standard deviation of the limiting Gamma distribution are given in Table~\ref{tab:prior-nu};
they encompass a wide range of different prior beliefs.
In this way the model becomes more stable, showing only minor variations when the 
limiting ranges of 
$\alpha$ and $\beta$ are modified.
It also becomes more flexible, and fits better the data as we show later in the experimental section.

\begin{table}[!h]
	\centering
	\begin{tabular}{@{}lrrrrr@{}}
		\toprule
		& \multicolumn{1}{l}{$\alpha$} & \multicolumn{1}{l}{$\beta$} & \multicolumn{1}{l}{mean} & \multicolumn{1}{l}{sd} & \multicolumn{1}{l}{p($\nu$ \textless 30)} \\
		\midrule
		\cite{kruschke2013bayesian} &                         2 &                      0.1 &                       20 &                     14 &                                   0.80 \\
		\cite{juarez2010model} &                         1 &                   0.0345 &                       29 &                     29 &                                   0.64 \\
		&                       0.5 &                     0.05 &                       10 &                     14 &                                   0.92 \\
		&                       0.5 &                     0.15 &                        3 &                      5 &                                   0.99 \\
		&                         5 &                     0.05 &                      100 &                     45 &                                   0.02 \\
		&                         5 &                     0.15 &                       33 &                     15 &                                   0.47 \\
		\bottomrule
	\end{tabular}
	\caption{Characteristics of the Gamma distribution for different values of $\alpha$ and $\beta$. The last four rows show the characteristic of the extreme distributions assumed by our hierarchical model.
	The hierarchical model however contains all the priors corresponding to intermediate values of  $\alpha$ and $\beta$.}
	\label{tab:prior-nu}
\end{table}

 The priors for the parameters of the high-level distribution are thus:
   \begin{align*}
   & \delta_0 \sim \mathrm{unif} (-1,1)  \\ 
   & \sigma_0 \sim \mathrm{unif}(0,\bar{\sigma_{0}}) \\
  & \nu \sim \mathrm{Gamma}(\alpha,\beta) \\
  & \alpha \sim \mathrm{unif}(\underline{\alpha},\bar{\alpha}) \\
  & \beta \sim \mathrm{unif}(\underline{\beta},\bar{\beta}) 
  \end{align*}
  
\subsection{The region of practical equivalence}

Our  knowledge about a parameter is fully represented by the posterior distribution. 
Yet it is handy to summarize the posterior in order to take decisions.
In  \citep{coraniML2015} we summarized the posterior distribution by reporting the probability of positiveness and negativeness;
however in this way we considered only the sign of the differences, neglecting their magnitude.

A more informative summary of the posterior is obtained introducing
a region of practical equivalence (rope), 
constituted by a range of parameter values that are practically equivalent to the null difference between the two classifiers.
We thus summarize the posterior distribution by reporting how much probability lies within the rope, at its left and at its right.
The limits of the rope are established by the analyst based on his experience; thus
there are no uniquely correct limits for  the rope \cite[Chap. 12]{kruschke2015doing}.

Dealing with the evaluation of classifiers, \cite{hand2006classifier} reports that the apparent superiority of 
a classifier in a simulation may be swamped by other sources of uncertainty when adopted in practice.
Examples of such sources of uncertainty are
the presence of non-stationary distributions or unintended biases
in the training samples. 
These  thin improvements measured in simulations might hardly imply any advantage in practice.

We thus consider two classifiers to be practically equivalent
if their mean difference of accuracy lies within (-0.01,0.01). 
This constitutes a  sensible default for the rope: yet a researcher could legitimately adopt a different rope.
If one is unsure about the limits of the rope, one could show
how the results vary for different rope widths. We provide an example in Sec.~\ref{sec:uci-signed-rank}.

The rope yields a realistic null hypothesis that can be verified. If a large mass of posterior probability
lies within the rope,  we claim the two classifiers to be practically equivalent. A sound approach to detect equivalent classifiers could be very useful in online model selection \citep{krueger2015fast} where one should quickly discard algorithms that perform the same.

\subsection{The inference of the test}

We focus on estimating the posterior distribution of the difference of accuracy between the two classifiers on a \textit{future unseen data set}. 
We compute the probability of left, rope and right being the most probable outcome 
on the next data set.

Thus we compute the probability by which
 $p(left)> \max(p(rope),p(right))$ or  $p(right)> \max(p(rope),p(left))$ or
 $p(rope)>\max(p(left),p(right))$. 
If one removes the rope,  this inference is similar to the inference performed by the  Bayesian sign-test.
\begin{figure}[!h]
	\centering
	\includegraphics[width=10cm]{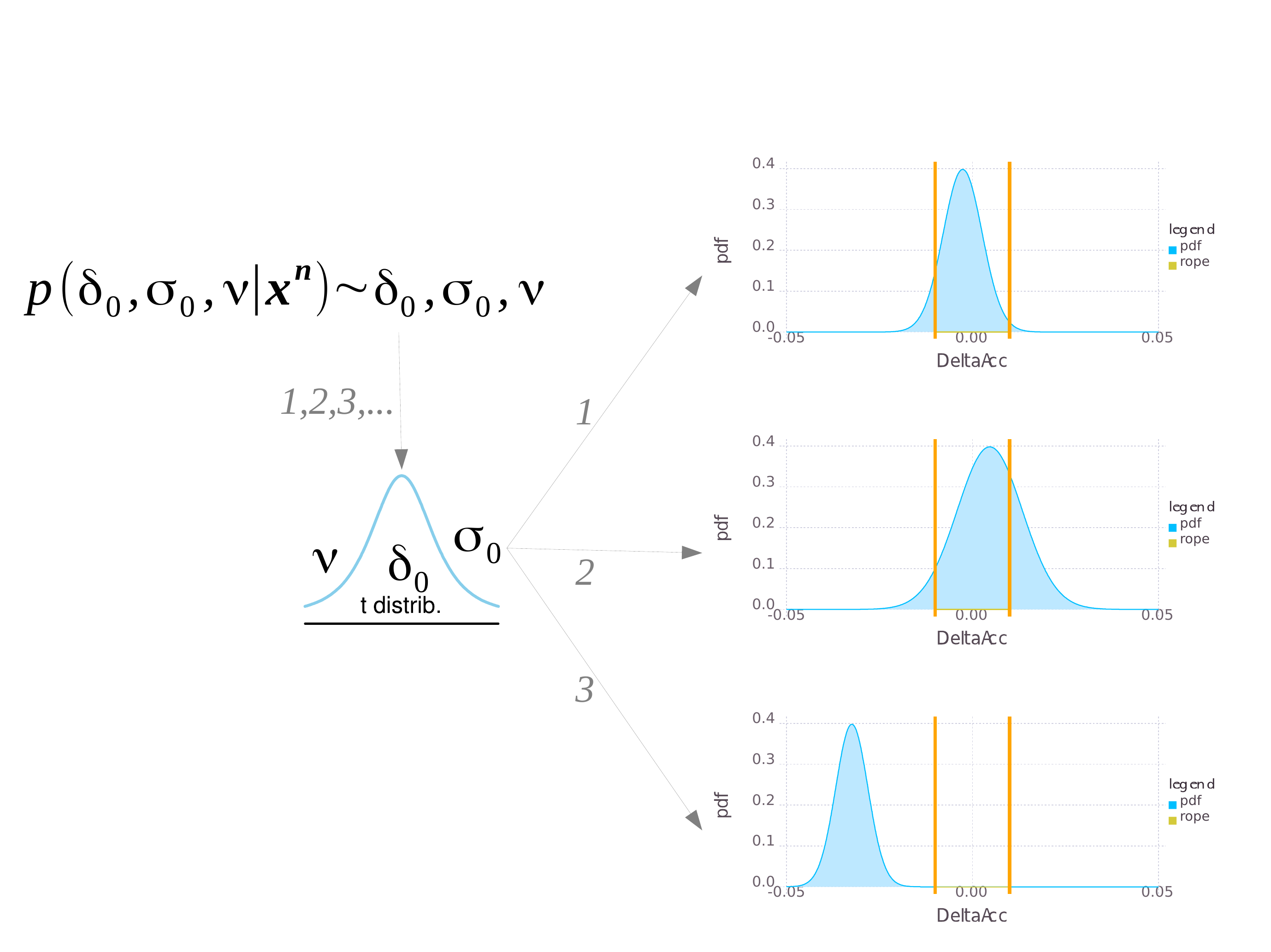}
	\caption{Diagram of the inference of the test.}
	\label{fig:infer}
\end{figure}

To compute such inference, we proceed  as follows:
\begin{enumerate}
	\item initialize the counters $n_{left}=n_{rope}=n_{right}=0$;
	\item for $i=1,2,3,\dots,N_s$ repeat
	\begin{itemize}
		\item sample $\mu_0, \sigma_0,\nu $ from the posterior of these parameters;
		\item  define the posterior of the mean difference accuracy on the next dataset, i.e., $t(\delta_{next};\delta_0, \sigma_0,\nu)$;
		\item from $t(\delta_{next};\delta_0, \sigma_0,\nu)$ compute the three probabilities $p(left)$ (integral on $(-\infty,r])$), $p(rope)$ (integral on $[-r,r]$) and $p(right)$ (integral on $[r,\infty)$);
		\item determine the highest among $p(left),p(rope),p(right)$ and  increment the respective counter $n_{left},n_{rope},n_{right}$;
	\end{itemize}
	\item compute  $P(left)=n_{left}/N_s$, $P(rope)=n_{rope}/N_s$ and $P(right)=n_{right}/N_s$;
	\item decision: when $P(rope)> 1-\alpha$ declare the two classifiers to be \textit{practically equivalent};
	when $P(left) >  1-\alpha$ or $P(right) >  1-\alpha$ we declare the two classifiers to be significantly different in the respective directions.  	
\end{enumerate}
We have chosen  $r=0.01$ and, thus, our region of practical equivalence (rope) is  $(-0.01, 0.01)$. 
Figure~\ref{fig:infer} shows a diagram of this inference schema and reports 
three sampled posteriors  $t(\delta_{next};\delta_0, \sigma_0,\nu)$. For these three cases we have that 
$(p(left),p(rope),p(right))$ are respectively (from top to bottom) $(0.08,0.90,0.02)$, $(0.05,0.67,0.28)$,  $(1,0,0)$
and so after these three steps $n_{left}=1$, $n_{rope}=2$, $n_{right}=0$ (in the next experiments we will consider $N_s=4000$).

\subsection{The shrinkage estimator for cross-validation}\label{sec:shrinkage}
The hierarchical model jointly estimates the $\delta_i$'s by applying
shrinkage to the $\bar{x}_i$'s.
In the uncorrelated case, the shrinkage estimator is known to be more accurate than the MLE.
In this section we show that the shrinkage estimator is more accurate than MLE also in the correlated
case, such as the data generated by cross-validation. This allows the hierarchical model to be more accurate than the existing method in the estimation of the $\delta_i$'s.

The $\delta_i$'s of the hierarchical model are independent given the parameters of the higher-level distribution.
If such parameters were known, the $\delta_i$'s would be conditionally independent and they would be independently estimated.
Instead such parameters are unknown, causing the $\delta_0$ and the $\delta_i$'s to be jointly estimated.
As a result
the estimate
of each $\delta_i$ is informed by data collected also on all the other data sets.
Intuitively, each data set informs the higher-level parameters, which in turn constrains and improves the parameters of the individual data sets \citep[Chap. 9]{kruschke2013bayesian}.

To show this, we assume the cross-validation results on the $q$ data sets to be generated by the hierarchical model:
\begin{align}
&	 \delta_i  \sim p(\delta_i), \nonumber \\
&	\mathbf{x}_{i} \sim MVN(\mathbf{1}\delta_i,\bm{\Sigma}). \label{eq:hier-simple}
\end{align}
where for simplicity we assumed the variances $\sigma_i^2$ of the individual data sets 
to be equal to $\sigma^2$ and known. Thus all data sets have the same covariance matrix $\bm{\Sigma}$, which is defined as follows: the  variances  equal $\sigma^2$ and the correlations equal $\rho$. 
Note that Eqn. (\ref{eq:hier-simple}) coincides with (\ref{eq:mvn}). This is a general model that makes no assumptions about the distribution $p(\delta_i)$.
We denote the 
first two moments of $p(\delta_i)$ as 
$E[\delta_i]=\delta_0$ and $\text{Var}[\delta_i]=\sigma_0^2$.

We study the MAP estimates of the parameters $\delta_1,\dots,\delta_m,\delta_o,\sigma_o^2$, which asymptotically tend to the Bayesian estimates. 
A hierarchical model is being fitted to the data. Such model is a simplified version of that presented in Sec.~\ref{sec:hier}. In particular
$p(\delta_i)$ is Gaussian for analytical tractability.
\begin{equation}
\label{eq:simplfiedmod0}
\begin{array}{l}
P(\bm{\bar{x}},\bm{\delta},\delta_0,\sigma_0^2)  =\prod\limits_{i=1}^{q} N(\mathbf{x}_{i};\mathbf{1}\delta_i,\bm{\Sigma}) N(\delta_i;\delta_o,\sigma_o^2)p(\delta_o,\sigma_o^2).\end{array}
\end{equation}
This model  is misspecified since $p(\delta_i)$ is generally not Gaussian.
Nevertheless, it correctly estimates 
the mean and variance of $p(\delta_i)$, as we show in the following.

\begin{Proposition} \label{prop:1}
	The derivatives of the logarithm of $P(\bm{\bar{x}},\bm{\delta},\delta_0,\sigma_0^2)$ are:
	\begin{align}
	\nonumber
	\frac{d}{d \delta_i}\ln(P(\cdot))&=\frac{\delta_o - \delta_i}{\sigma_o^2} + \frac{\bar{x}_i-\delta_i}{\sigma_n^2},\\
	\nonumber
	\frac{d}{d \delta_o}\ln(P(\cdot))&=\frac{-q \delta_o + \sum\limits_{i=1}^q \delta_i}{ \sigma_o^2}+\frac{d}{d\delta_o}\ln(p(\delta_o,\sigma_o^2)),\\
	\nonumber
	\frac{d}{d \sigma_o}\ln(P(\cdot))&=\frac{q \delta_o^2 + \sum\limits_{i=1}^q \delta_i^2  - 2 \delta_o \sum\limits_{i=1}^q \delta_i - q \sigma_o^2}{\sigma_o^3}+\frac{d}{d\sigma_o}\ln(p(\delta_o,\sigma_o^2)).
	\end{align}
\end{Proposition}
If we further assume that $p(\delta_o,\sigma_o^2) \approx \text{constant}$ (flat prior), by equating the derivatives to zero, 
we derive the following consistent estimators:
\begin{align}
\sigma_o^2 =\frac{1}{q}\sum\limits_{i=1}^q (\hat{\delta}_i-\hat{\delta}_o)^2, \\
&  \hat{\delta}_i=\frac{\hat{\sigma}_o^2 \bar{x}_i + \sigma_n^2 \tfrac{1}{q}\sum\limits_{i=1}^q \bar{x}_i }{\hat{\sigma}_o^2 + \sigma_n^2}=w\bar{x}_i  + (1-w)\tfrac{1}{q}\sum\limits_{i=1}^q \bar{x}_i,
\end{align}
where $w=\hat{\sigma}_o^2/(\hat{\sigma}_o^2+\sigma_n^2)$ and, to keep a simple notation, we have not explicited the expression $ \hat{\sigma}_o $ as a function of $\bar{x}_i,\sigma_n^2$. 
Notice that the estimator $\hat{\delta}_i$ 
shrinks the estimate towards $\tfrac{1}{q}\sum_{i=1}^q \bar{x}_i$ that is an estimate of $\delta_0$. 
Hence, the Bayesian hierarchical model consistently estimates  $\delta_0$ and $\sigma_0^2$ from data and converges to the shrinkage estimator $\hat{\delta}_i(\mathbf{x}_i)=w\bar{x}_i+(1-w)\delta_0$.

It is known that the shrinkage estimator achieves a lower error than MLE in case of uncorrelated data;
see \citep[Sec 6.3.3.2]{murphy2012machine} and the references therein.
However there is currently no analysis of  shrinkage with correlated data, such as those yielded by cross-validation.
We study this problem in the following.

Consider the generative model (\ref{eq:hier-simple}). The likelihood regarding the i-th data set is:
\begin{equation}
\begin{array}{l}
p(\mathbf{x}_i|\delta_i,\bm{\Sigma})=N(\mathbf{x}_i;\mathbf{1}\delta_i,\bm{\Sigma})   =\dfrac{\exp(-\frac{1}{2}(\mathbf{x}_i-\mathbf{1}\delta_i)^{T}\bm{\Sigma}^{-1}(\mathbf{x}_i-\mathbf{1}\delta_i))}{(2\pi)^{n/2}\sqrt{|\bm{\Sigma}|}}.
\label{eq:lik-correlated}\vspace{2mm}\\
\end{array}
\end{equation}
Let us denote by $\bm{\delta}$ the vector of the $\delta_i$'s.
The joint probability of data and parameters is:
$$
P(\bm{\delta},\mathbf{x}_1,\ldots,\mathbf{x}_q)= \prod_{i=1}^{q} N(\mathbf{x}_i;\mathbf{1}\delta_i,\bm{\Sigma})p(\delta_i).
$$
Let us focus on the i-th group, denoting by $\hat{\delta}_i(\mathbf{x}_i)$ an estimator of $\delta_i$.
The mean squared error (MSE) of the estimator w.r.t. the true joint model
$P(\delta_i,\mathbf{x}_i)$ is:
\begin{equation}
\label{eq:mese}
\iint \left(\delta_i-\hat{\delta}_i(\mathbf{x}_i)\right)^2 N(\mathbf{x}_i;\mathbf{1}\delta_i,\bm{\Sigma})p(\delta_i)d\mathbf{x}_i d\delta_i.
\end{equation}   
\begin{Proposition} \label{prop:2}
	The MSE of the maximum likelihood estimator is:
	\begin{align}
	\nonumber
	\mathrm{MSE_{MLE}}&=\iint \left(\delta_i-\bar{x}_i\right)^2 N(\mathbf{x}_i;\mathbf{1}\delta_i,\bm{\Sigma})p(\delta_i) d\mathbf{x}_i d\delta_i\\
	\nonumber
	&=\frac{1}{n^2}\mathbf{1}^T\bm{\Sigma}\mathbf{1},
	\end{align}
	which we denote in the following
	also as $\sigma_n^2=\frac{1}{n^2}\mathbf{1}^T\bm{\Sigma}\mathbf{1}$.
\end{Proposition}
Now consider the shrinkage  estimator $\hat{\delta}_i(\mathbf{x}_i)=w\bar{x}_i+(1-w)\delta_0$ with $w \in (0,1)$, which pulls the MLE estimate $\bar{x}_i$ towards the mean $\delta_0$ of the upper-level distribution. 
\begin{Proposition} \label{prop:3}
	The MSE of the shrinkage estimator is:
	\begin{align}
	\nonumber
	\mathrm{MSE_{SHR}} &=\iint \left(\delta_i-w\bar{x}_i-(1-w)\delta_0\right)^2 N(\mathbf{x}_{i};\mathbf{1}\delta_i,\bm{\Sigma})p(\delta_i) d\mathbf{x}_{i} d\delta_i\\
	\nonumber
	&=w^2\sigma_n^2+ (1-w)^2\sigma_0^2.
	\end{align}
\end{Proposition}
As we have seen, the hierarchical model converges to the shrinkage estimator with $w=\sigma_0^2/(\sigma_0^2+\sigma_n^2)$. Then: 
\begin{align}
\nonumber
\mathrm{MSE_{SHR}}&=w^2\sigma_n^2+ (1-w)^2\sigma_0^2
=\frac{\sigma_0^4+\sigma_n^2\sigma_0^2}{(\sigma_0^2+\sigma_n^2)^2}\sigma_n^2\\
\nonumber
&= 
\frac{\sigma_0^2}{(\sigma_0^2+\sigma_n^2)}\sigma_n^2<\sigma_n^2=\mathrm{MSE_{MLE}}.
\label{eq:mle-shrink}
\end{align}

\textit{Therefore, the shrinkage estimator achieves a smaller mean squared error 
	than the MLE.
}

\subsection{Implementation and code availability}
  We implemented the hierarchical model in Stan \citep{carpenterstan}, a language for Bayesian inference.  
  In order to improve the computational efficiency, we 
  exploit a quadratic matrix form to compute simultaneously the 
  likelihood of  the $q$ data sets.
This provides a speedup of about one order of
magnitude compared to the naive implementation in which the likelihoods are computed separately on each data set.
  Inferring the hierarhical model on the results of 10 runs of 10-folds cross-validation on 50 data sets (a total of 5000 observations) takes about three minutes on a standard laptop.
  
	 The Stan code is available from \url{https://github.com/BayesianTestsML/tutorial/tree/master/hierarchical}.
The same repository provides the R code of all the simulations of Sec.~\ref{sec:experiments}.

  \section{Experiments}\label{sec:experiments}
  \subsection{Estimation of the $\delta_i$'s under misspecification of p($\delta_i$)}
According to the proofs of Sec.~\ref{sec:hier},
 the shrinkage estimator of the $\delta_i$'s
has lower mean squared error 
than the maximum likelihood estimator, constituted by the arithmetic mean of the 
cross-validation results.
This result holds
even if the $p(\delta_i)$ of the hierarchical model is misspecified: it
only requires the hierarchical model to 
reliably estimate the first two moments of $p(\delta_i)$.

To verify this theoretical result we design the following experiment.
    We consider these numbers of data sets: $q=\{5,10,50\}$.
  For each value of $q$ we repeat 500 experiments consisting of:
  \begin{itemize}
	  \item  sampling of the $\delta_i$'s ($\delta_1, \delta_2,\ldots,\delta_q$) from the \textit{bimodal mixture}
$$p(\delta_i)= \pi_1 N(\delta_i|\mu_1,\sigma_1)
+ \pi_2 N(\delta_i|\mu_2,\sigma_2)
$$ with $k$=2,
 $\mu_1$=0.005, $\mu_2$=0.02, $\sigma_1$=$\sigma_2$=$\sigma$=0.001, $\pi_1=\pi_2=0.5$.
  	
  	\item For each $\delta_i$:
  	\begin{itemize}
  	 \item implement two classifiers 
  	whose actual difference of accuracy is $\delta_i$, following the procedure
  	given in Appendix; 
  	\item perform 10 runs of 10-folds cross-validation with the two classifiers; 
  	\item measure the mean of the cross-validation results $\bar{x}_i$ (MLE).
  	\end{itemize}
	 \item Infer the hierarchical model using the results referring to the $q$ data sets;
	 \item obtain the shrinkage estimates of each $\delta_i$;
  	\item measure $\mathrm{MSE_{MLE}}$ and $\mathrm{MSE_{SHR}}$ as defined in Sec.~\ref{sec:shrinkage}.
  \end{itemize}

\begin{table}[!ht]
	\rowcolors{2}{lightblue}{white}
	\centering
	\begin{tabular}{@{}ccc}
		\toprule
		$q$ & \multicolumn{2}{c}{\textit{Mean Squared Error}} \\
		& $\mathrm{MLE}$   & $\mathrm{Shrinkage}$ \\ \midrule
		5 & .00036 & .00017 \\
		10 & .00036 & .00014 \\
		50 & .00036 & .00012 \\ 
		\bottomrule
	\end{tabular}
	\caption{Estimation error of the $\delta_i$'s. The scale of the actual errors on the estimation of the $\delta_i$'s
		can be realized considering that for instance 0.02$^2$=.0004.
		\label{tab:MSE}}
\end{table}

As reported  in Tab.~\ref{tab:MSE}, $\mathrm{MSE_{SHR}}$ is generally 50\% lower than
  $\mathrm{MSE_{MLE}}$ for every value of $q$. This verifies our theoretical findings. 
  It also shows that the mean of the cross-validation estimates is a quite noisy estimator of $\delta_i$,
  even if 10 repetitions of cross-validation are performed. The problem is that all such results are correlated and thus they have limited informative content.

Interestingly, the MSE of the shrinkage estimator
decreases with 
  $q$. Thus the presence of more data sets allows
  to better estimate the moments of $p(\delta_i)$, improving the shrinkage estimates as well.
  Instead the error of the MLE does not vary with $q$ since the parameters of each data set are independently estimated.

In the uncorrelated case it is well-known that, on average, the shrunken estimate
is much closer to the true parameters than the MLE is. Our results extend this finding to the correlated case.

      \subsection{Comparison of equivalent classifiers}\label{sec:sim-null}
      In this section we adopt a Cauchy distribution as $p(\delta_i)$; this is an idealized situation in which the hierarchical model can recover the actual $p(\delta_i)$. 
      We will relax this assumption in  Sec.~\ref{sec:friedman}.
      
      We simulate
      the null hypothesis of the signed-rank setting the median of the Cauchy to $\delta_0=0$. 
      We set the scale factor of the distribution to 
	  1/6 of the rope length; this implies that 80\% of the sampled   $\delta_i$'s lies within the rope,
	  which is by far the most probable outcome.

      We consider the following numbers of data sets: $q=\{10,20,30,40,50\}$.
      For each value of $q$ we repeat 500 experiments consisting of:
      \begin{itemize}
       \item sampling the $\delta_i$'s ($\delta_1, \delta_2, \ldots, \delta_q$) from $p(\delta_i)$;
         	\item for each $\delta_i$:
         	\begin{itemize}
         		\item implement two classifiers 
         		whose actual difference of accuracy is $\delta_i$, following the procedure
         		given in Appendix; 
         		\item perform 10 runs of 10-fold cross-validation with the two classifiers; 
         	\end{itemize}
       \item analyze the results through the signed-rank and the hierarchical model.
      \end{itemize}

		\begin{figure}[!ht]
			\centering
			\includegraphics[width=15cm]{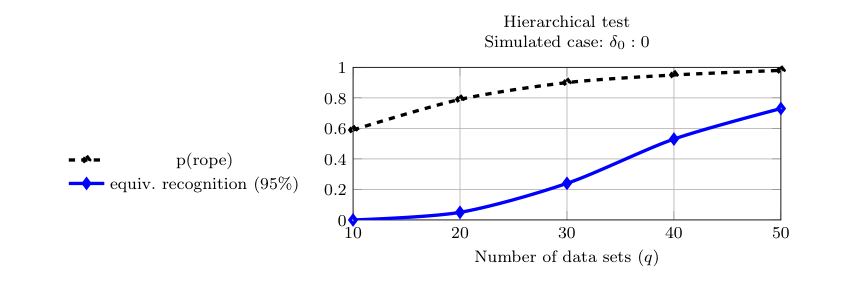}
			\caption{Behavior of the hierarchical classifier when dealing with two  equivalent classifiers.
				\label{fig:rope-recognition}
			}
		\end{figure}
		
      The signed-rank test ($\alpha$=0.05) rejects the null hypothesis about 5\% of the times
      for each value of $q$. 
      It is  thus correctly calibrated.
Yet, it provides no valuable insights.
      When it does not reject $H_0$ (95\% of the times), it does \textit{not} allow claiming that the null hypothesis is true. 
	When it rejects the null (5\% of the times), it draws a \textit{wrong} conclusion since $\delta_0$=0.

        The hierarchical model draws more sensible conclusions. 
The posterior probability $p(rope)$ increases with $q$ (Fig.~\ref{fig:rope-recognition}): 
the presence of more data sets 
provides more evidence that they are equivalent.
        For $q$=50 (the typical size of a machine learning study), the average p(rope) reported in simulations 
        is larger than 90\%.
        Fig.~\ref{fig:rope-recognition} reports also 
        on \textit{equivalence recognition}, which is the proportion of simulations  
        in which p(rope) exceeds 95\%.   Equivalence recognition increases with $q$, reaching about 0.7 for $q$=50.

      Moreover in our simulations the hierarchical model
      never estimated p(left)\textgreater 95\% or p(right)\textgreater 95\%, so it made no Type I errors. 
      In fact nsht is tied to commit a 5\% Type I error when the null hypothesis is true; but
      this is not the case of Bayesian estimation with rope, which instead generally makes less Type I errors \citep{kruschke2013bayesian}
        than nhst. 
        As we described earlier, this is at least in part a consequence of using a whole interval instead of just a single point for the null-hypothesis, which makes the null-hypothesis more difficult to reject.\\

        \textbf{Running the signed-rank twice?}
        We \textit{cannot} detect practically equivalent classifiers by running twice the signed-rank test, 
        e.g., once with null hypothesis $\delta_0=$0.01 and once with the null hypothesis $\delta_0 =$-0.01.
        Even if the signed-rank test does not reject the null in both cases, we still cannot affirm that the two classifiers 
        are equivalent, since non-rejection of the null does not allow claiming that the null is true.

      \subsection{Comparison of practically equivalent classifiers}
      
      We now simulate two classifiers whose actual difference of accuracy is practically irrelevant but different from zero.
      We consider
      two classifiers whose average difference is $\delta_0$=0.005, thus within the rope.
      
	  We consider $q=\{10,20,30,40,50\}$.
      For each value of $q$ we repeat 500 experiments as follows:
      \begin{itemize}
      \item set $p(\delta_i)$ as a Cauchy distribution with $\delta_0$=0.005 and the same scale factor as in previous experiments (the rope remains by far the most probable outcome for the sampled $\delta_i$'s);
       \item sample the $\delta_i$'s ($\delta_1, \delta_2, \ldots, \delta_q$) from $p(\delta_i)$;
       \item   implement for each $\delta_i$ two classifiers whose actual difference of accuracy is
       $\delta_i$ and perform 10 runs of 10-fold cross-validation; 
       \item analyze the cross-validation results through the signed-rank and the hierarchical model.
      \end{itemize}

      \begin{figure}[!ht]
      	\centering
      	\includegraphics[width=15cm]{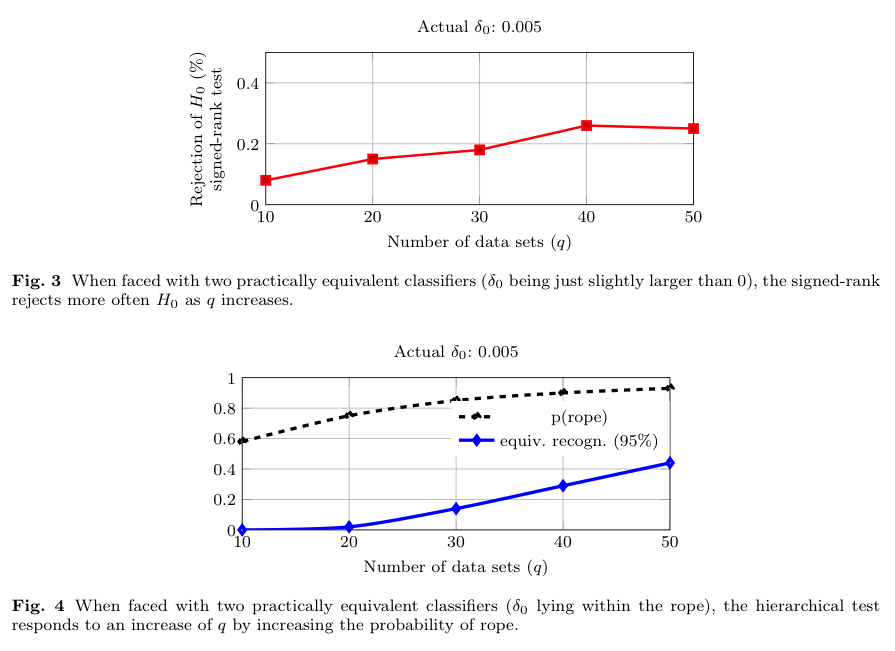}
      \end{figure}

      The signed-ranked test is more likely to reject the null hypothesis as the number of data sets increases (Fig.~3). 
       When 50 data sets are available, the signed-rank rejects the null
       in about 25\% of the simulations, despite the trivial difference between the two classifiers.
      Indeed one can reject the null of the signed-rank test when comparing two almost equivalent classifiers, by comparing them on enough data sets. 
      As reported in the ASA statement on p-value \citep{wasserstein2016asa}
      even a tiny effect  can produce a small p-value
if the sample size is large
enough.
                    
The behavior of the hierarchical test is the opposite and more sensible. With increasing number of data sets on which the classifiers show similar performance, the researcher should be more convinced that the two classifiers are practically equivalent. The hierarchical test indeed increases the posterior probability of rope (Fig.~4). 
It is slightly less effective in recognizing equivalence  than in the previous experiment since $\delta_0$ is now closer to the limit of the rope.
When $q$=50, it declares equivalence detection with 95\% confidence in about 40\% of the simulated cases.
In our simulations it never claims a significant difference, namely it never
 estimates 
p(left) or p(right) to be larger than 95\%.
  
The hierarchical test  is thus effective 
at detecting classifiers that are (actually or practically) equivalent.
This is impossible for the signed-rank test: it gives the correct answer that the classifiers are different, but in terms of the single-point hypothesis. On the contrary, the hierarchical test claims that they are the same in terms of the interval-based hypothesis. 

Moreover the signed-rank ($\alpha$=0.05)
is tied to commit 5\% Type I errors  
under the null hypothesis, even in presence of infinite data.
The hierarchical model is free from this constraint and in our simulation it made no
Type I errors at all.
 
The hierarchical model  is thus more conservative than the signed rank test.
The price to be paid is that it might be less powerful 
at
claiming significance when comparing two classifiers whose accuracies are truly different.
We investigate this setting in the next section.

       \subsection{Simulation of practically different classifiers}
      We now simulate two classifiers which are significantly different.
      We consider different values of $\delta_0$: $\{0.015, 0.02, 0.025, 0.03\}$. 
      We set the scale factor of the Cauchy to $\sigma_0$=0.01 and     
      the number of data sets to $q$=50.
      
      We repeat 500 experiments for each value of $\delta_0$, organized as in the previous sections.
       We then check the \textit{power} of the two tests for each value of $\delta_0$.
      The power of the signed-rank is the proportion of simulations in which it rejects the null hypothesis ($\alpha$=0.05).
      The power of the hierarchical test is the proportion of simulations in which it estimate p(right)\textgreater 0.95.
      
          The hierarchical model is necessarily less powerful than the signed-rank: to declare the two classifiers as significantly different, the signed-rank has to reject the null hypothesis $\delta_0=0$, while the hierarchical model needs to have a  posterior mass larger than 0.95 in the region to the right of the rope.
          As shown in Fig.~5, the signed-rank test is indeed more powerful, especially  when $\delta_0$ lies  just slightly outside the rope.
      The two tests show similar power when $\delta_0$ is larger than 0.02.

\begin{figure}[!ht]
	\centering
	\includegraphics[width=15cm]{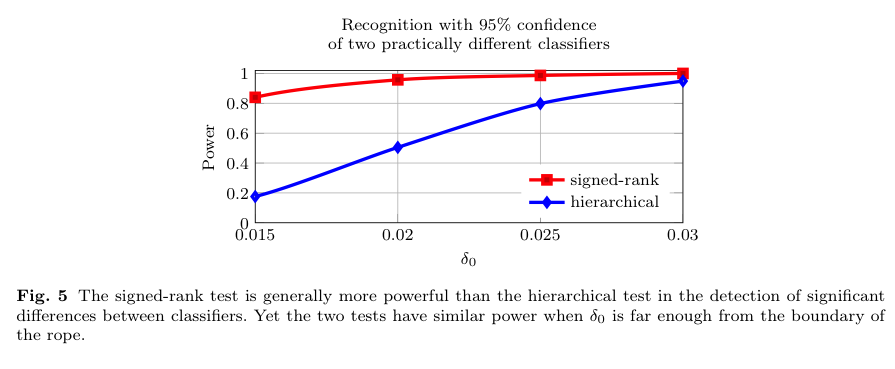}
\end{figure}

\subsection{Discussion}
The main experimental findings so far are as follows.
First, the shrinkage  estimator of the $\delta_i$'s 
yields a lower mean squared error than the
MLE estimator, even under misspecification of $p(\delta_i)$.

Second,
the hierarchical model is effective at detecting equivalent classifiers;
this is instead impossible for the nhst test.

Third, the hierarchical model
is more conservative than the signed-rank test: it rejects more rarely the null hypothesis, as a consequence of its null hypothesis being more realistic.
For this reason
it commits less Type I error than the signed-rank.

However,
it is less powerful than the signed-rank:
when comparing two  significantly different classifiers, 
the hierarchical test  claims 95\% significance less frequently than the signed-rank.
The difference in power is not necessarily a large one, as shown in the previous simulation.

Moreover the probabilities returned by the hierarchical model can be interpreted
in a more flexible way than simply checking if there is an outcome whose
posterior probability is larger than 95\%,
as we discuss in the next section.

\subsection{Interpreting posterior odds}
The ratio of posterior probabilities (\textit{posterior odds}) shows
the extent to which the data support one hypothesis over the other.
For instance we can compare the support for left and right by
computing the posterior odds $o(\mathrm{left,right})=\frac{p(left)}{p(right)}$.
When $o(\mathrm{left,right}) > 1$ there is evidence in favor of left;
when $o(\mathrm{left,right}) < 1$ there is evidence in favor of right.
Rules of thumb for interpreting the amount of evidence 
corresponding to posterior odds are discussed by \cite{raftery1995bayesian} 
and summarized in Tab.~\ref{tab:odds}:

     \begin{table}[!ht]
\rowcolors{1}{white}{lightblue}
     	\centering
     	\begin{tabular}{@{}cc@{}}
     		\toprule
     		Posterior Odds           & Evidence \\ \midrule
     		1--3           & weak     \\
     		3--20          & positive \\
     		\textgreater20 & strong   \\ \bottomrule
     	\end{tabular}
     	\caption{Grades of evidence corresponding to posterior odds.}
     	\label{tab:odds}
     \end{table}
      
Thus even if none of the three probabilities exceeds the 95\% threshold, 
we can still draw meaningful conclusions by
interpreting the posterior odds. We will adopt this approach in the following simulations.

The p-values  cannot be interpreted in a similar fashion, since they are affected
both by sample size and effect size.
In particular \citep{wasserstein2016asa} show that
smaller p-values do not necessarily imply the presence of larger effects and larger p-values do not imply a lack of effect. A tiny effect can produce a small p-value if the sample size is large enough, and large effects may produce unimpressive p-values if the sample size is small.

\subsection{Experiments with Friedman's functions}\label{sec:friedman}
The results presented in the previous sections refer to conditions in which the actual $p(\delta_i)$ (misspecified or not) is analytically known.
In this section we perform experiments in which the $\delta_i$'s  are not sampled from an analytical distribution; rather, they are due to different settings of sample size, noise etc. 
The actual $p(\delta_i)$ of the next section has no analytical form and thus the $p(\delta_i)$
of the hierarchical model is unavoidably misspecified. This is a challenging setting for checking the conclusion of the hierarchical model.

We generate
data sets via
the three functions ($F\#1$,
$F\#2$ and $F\#3$) proposed by \cite{friedman1991multivariate}.

Function $F\#1$ contains ten features $x_1,\ldots,x_{10}$, each uniformly distributed over $[0,1]$. Only five features are used to generate the response $y$:
\begin{align*}
F\#1: \,\,	y = 10 sin(\pi x_1 x_2) + 20 (x_3 - 0.5)^2 + 10 x_4 + 5 x_5 + \epsilon_1,
\end{align*}
where $\epsilon_1 \sim N(0,1)$.
To turn this regression problem into a classification problem, we discretize $y$ in two bins.
The two bins are separated by the median of $y$, which we independently estimate on a sample of  10,000 generated instances.

Functions $F\#2$ and $F\#3$ have four features $x_1,\ldots,x_{4}$ uniformly distributed over the ranges:
\begin{align*}
0 \leq x_1 \leq 100, \\
40 \pi \leq x_2 \leq 560 \pi, \\
0 \leq x_3 \leq 1, \\
1 \leq x_4 \leq 11. \\
\end{align*}

The functions are:
\begin{align*}
F\#2: \,\,	y = (x_1^2 + (x_2 x_3 - (1/x_2 x_4))^2)^{0.5} + \epsilon_2 \\
F\#3: \,\,	y = \arctan  \left( \frac{x_2 x_3 - (1/x_2x_4)}{x_1} \right) + \epsilon_3 
\end{align*}
where $\epsilon_2 \sim N(0,\sigma_{\epsilon_2}^2)$
and $\epsilon_3  \sim N(0,\sigma_{\epsilon_3}^2)$. The original paper sets $\sigma_{\epsilon_2}$=125
and $\sigma_{\epsilon_3}$=0.1. Also in this case we turn the problems from regression to classification by 
discretizing the response variable in two bins, delimited by its median.

We consider 18 settings for each function, obtained by varying the sample size ($n$) and the standard deviation of the noise (considering also twice and half the original values).
As a further factor we either consider only the original features or we add further twenty normally distributed random features.
We have overall 54 settings: 18 settings for each function. They are
summarized in Table~\ref{tab:friedman-settings}.

\begin{table}[!h]
\centering
\rowcolors{1}{white}{lightblue}
\begin{tabular}{@{}ccccc@{}}
\toprule
Function  type & $\sigma_{\epsilon}$ & $n$             & random Feats & Tot settings              \\
\midrule
F\#1              & \{0.5,1,2\}         & \{30,100,1000\} & \{0,20\}           & 3 $\cdot$ 3 $\cdot$ 2 =18 \\
F\#2              & \{62.5,125,250\}    & \{30,100,1000\} & \{0,20\}           & 3 $\cdot$ 3 $\cdot$ 2 =18 \\
F\#3              & \{0.05,0.1,0.2\}    & \{30,100,1000\} & \{0,20\}           & 3 $\cdot$ 3 $\cdot$ 2 =18 \\
\bottomrule
\end{tabular}
\caption{Settings used for generating data
with the Friedman functions.}
\label{tab:friedman-settings}
\end{table}

As a pair of classifiers we consider linear discriminant analysis (\textit{lda}) and classification trees (\textit{cart}), as implemented in the \texttt{caret} package for R, without any hyper-parameter tuning.
As first step we need to 
measure the actual $\delta_i$ between two given classifiers in each setting, which then allows us to know
the population of the $\delta_i$'s.

Our second step will be to check 
the conclusions of the signed-rank test and of the 
hierarchical model when they are provided with cross-validation results referring 
to a subset of settings.

\subsubsection*{Measuring $\delta_i$}
First we need 
to measure
the actual difference of accuracy $\delta_i$
between \textit{lda} and \textit{cart} in the i-th setting.
Taking advantage of the synthetic nature of the data, we adopt  the following procedure:
\begin{itemize}
\item for j=1:500
\begin{itemize}
\item  sample training data according to the specifics of the $i$-th setting: \textless function type, $n$, $\sigma_{\epsilon}$, number of random features \textgreater;
\item  fit \textit{lda} and \textit{cart} on the generated training data; 
\item  sample a large test set (5000 instances) and measure the difference of accuracy $d_{ij}$ between \textit{cart} and \textit{lda};
\end{itemize}
\item set $\delta_i \simeq 1/500 \sum_j d_{ij}$ .
\end{itemize}

Our procedure yields accurate estimates since each repetition 
is provided with independent data and has available a large test set.

For instance  in a certain setting two classifiers have mean difference of accuracy $\bar{x}$=0.09,
with standard deviation $s$=0.06.
The 95\% confidence interval of their difference is tight:
\begin{align*}
& \bar{x} \pm  1.96 \cdot \frac{s}{\sqrt{n}} = 
 0.09 \pm  1.96 \cdot \frac{0.06}{\sqrt{500}} = [0.085 - 0.095].
\end{align*}

If instead we had performed 500 runs of 10-folds cross-validation obtaining the same value of $\bar{x}$ and $s$,
the confidence interval of our estimates would be about 3.5 times larger,
as the standard error 
would be $s\sqrt{\frac{1}{n}+\frac{\rho}{1-\rho}}$ instead of ${\frac{s}{\sqrt{n}}}$, as shown in Eqn.(\ref{eq:ttest-nadeau}).

\subsubsection*{Ground-truth}
We compute the $\delta_i$ of each setting using the above procedure.
The ground-truth is that lda is significantly more accurate than cart.
More in detail, 65\% 
of the $\delta_i$'s belong to the region to the right of the rope (lda being significantly more accurate than cart).
Thus right is is the most probable outcome of the next $\delta_i$.
Moreover, the mean of the $\delta_i$'s is $\delta_0$=0.02 in favor of lda. 
\\

\subsubsection*{Assessing the conclusions of the tests}
We 
run 200  times the following procedure:
\begin{itemize}
 \item random selection 12 out of 18 settings for each Friedman function,
 thus selecting 36 settings;
 \item in each setting:
 \begin{itemize}
  \item  generate a data set according to the specific of the setting;
  \item run 10 runs of 10-folds cross-validation of lda and cart using paired folds;
 \end{itemize}
\item analyze the cross-validation results on the $q$=36 data sets using the signed rank and the hierarchical test.
\end{itemize}

We start checking the \textit{power} of the tests,
defined as the proportion of simulations in which the
null hypothesis is rejected (signed-rank) or the posterior probability p(right)
exceeds 95\% (hierarchical test).

The two tests have roughly the same power:
28\% for the signed-rank and 
27.5\% for the hierarchical test.
In the remaining simulations the signed-rank does not reject $H_0$; in those cases it
conveys no information since the p-values cannot be interpreted. 

We can instead interpret the posterior odds yielded by the hierarchical model, obtaining the following results:
\begin{itemize}
	\item in 11\% of the simulations both $o(right,rope)$
	and $o(right,left)$ are larger than 20, providing strong 
	evidence in favor of lda even if p(right) does not exceed 95\%;
	\item in a further 33\% of the simulations both $o(right,rope)$
	and $o(right,left)$ are larger than 3, providing
	at least positive evidence in favor of lda.
\end{itemize}

We have moreover to point out 
a 2\% of simulations in which the posterior odds provide erroneously positive evidence for rope over both right and left.
In no case there is positive evidence for left over either rope or right.
	
Thus the interpretation of posterior odds  allows drawing meaningful conclusions 
even when the 95\% threshold is not exceeded.
The probabilities are sensibly estimated, even if $p(\delta_i)$ is unavoidably misspecified.

\begin{figure}[!ht]
	\centering
	\includegraphics[width=15cm]{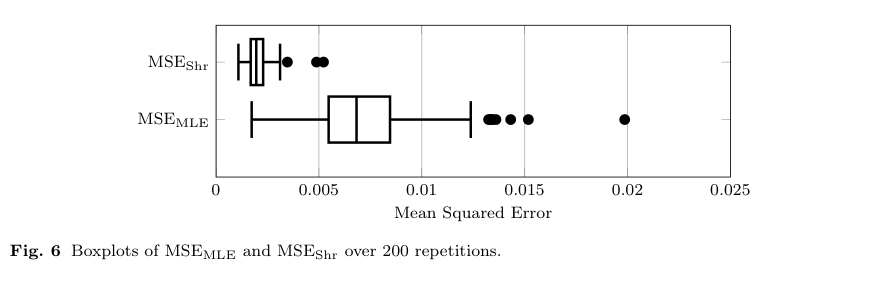}
\end{figure}

As a further check we compare 
$\mathrm{MSE_{MLE}}$ and 
$\mathrm{MSE_{Shr}}$.
Also in this case $\mathrm{MSE_{MLE}}$ is much lower than
$\mathrm{MSE_{Shr}}$ (Fig~6  ), with an 
average reduction of about  60\%.
This further confirms the properties of the shrinkage estimator.

\subsection{Sensitivity analysis on real-world data sets}\label{sec:real}
We now consider real data sets.
In this case we \textit{cannot} know the actual $\delta_i$'s:
we could repeat a few hundred times cross-validation but 
the resulting estimates
would have large uncertainty as already discussed.

We exploit this setting to perform sensitivity analysis and to further
compare the conclusions drawn by the hierarchical model and of the signed-rank test.

We consider 54 data sets taken from the WEKA data sets page\footnote{\url{http://www.cs.waikato.ac.nz/ml/weka/datasets.html}}. 
We consider five classifiers: naive Bayes (nbc), averaged-one dependence estimator (aode), hidden naive Bayes (hnb), decision tree (j48), grafted decision tree (j48gr). \cite{witten2011data} provides a summary description of all such classifiers with pointers to the relevant papers.
We perform 10 runs of 10-folds cross-validation for each classifier on each data set.
We run all experiments  using the WEKA\footnote{\url{http://www.cs.waikato.ac.nz/ml/weka/}} software.

A fundamental step of Bayesian analysis is to check how the posterior conclusions depend on the chosen prior and how the model fits the data. 
The hierarchical model shows some sensitivity on the choice of $p(\delta_i)$, 
being instead robust to the other assumptions (see later for further discussion).
The Student distribution 
is more flexible than the Gaussian and we have found that it consistently provides better fit to the data.
Yet, the model conclusions are sometimes sensitive on 
the prior on the degrees of freedom $p(\nu)$ of the Student. 
  
  In Table~\ref{tab:posterior-hier-jua} we compare the posterior inferences of the model, using the prior  $p(\nu) = Gamma(2,0.1)$ (proposed in a different context by \cite{juarez2010model}) or using the 
  more flexible model described in Sec.~\ref{sec:hier}, where the
  the parameters of the Gamma are described as further random variables with their own prior distributions.
  Such two variants are referred to as \textit{Gamma(2,0.1)} and \textit{hierarchical} in Table~\ref{tab:posterior-hier-jua}.
   
   \begin{table}[!ht]
   	\rowcolors{3}{white}{lightblue}
   	\centering
   	\begin{tabular}{@{}lllllll@{}}
   		\toprule
   		& \multicolumn{3}{c}{\textbf{Hierarchical}} & \multicolumn{3}{c}{\textbf{Gamma(2,0.1)}} \\ 
   		pair       & left   & rope   & right  & left   & rope  & right  \\ \midrule
   		nbc-hnb    & 1.00 & 0.00 & 0.00 & 1.00 & 0.00 & 0.00 \\
   		nbc-j48    & 0.80 & 0.02 & 0.18 & 0.80 & 0.01 & 0.20 \\
   		nbc-j48gr  & 0.84 & 0.02 & 0.14 & 0.84 & 0.01 & 0.15 \\
   		hnb-j48    & 0.03 & 0.10 & 0.87 & 0.03 & 0.02 & 0.95 \\
   		hnb-j48gr  & 0.03 & 0.07 & 0.90 & 0.03 & 0.02 & 0.95 \\
   		j48-j48gr  & 0.00 & 1.00 & 0.00 & 0.00 & 1.00 & 0.00 \\
   		\bottomrule
   	\end{tabular}
   	\caption{Posterior probabilities computed by two variants of the hierarchical model.}
   	\label{tab:posterior-hier-jua}
   \end{table}  
  In some cases the estimates of the two models differ by some points (Tab.~\ref{tab:posterior-hier-jua}). This means that the actual high-level distribution from which the $\delta_i$'s are sampled is not a Student (or a Gaussian), otherwise the estimate of the two models would converge.
  
  \begin{figure}[!ht]
  	\centering
  	\includegraphics[width=15cm]{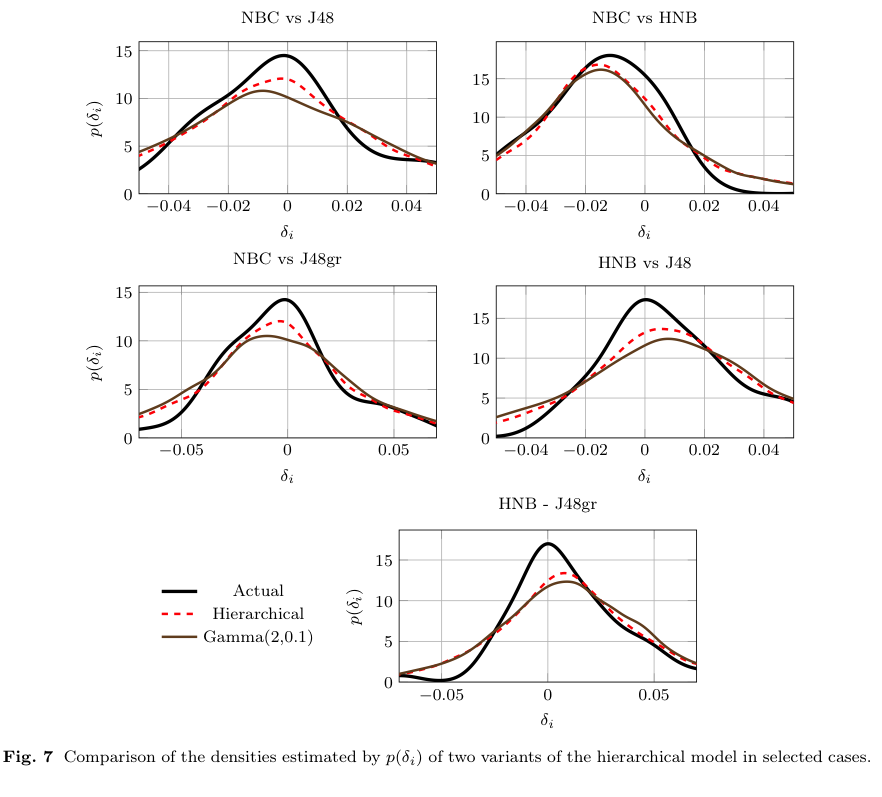}
  \end{figure}
  
  Which model better fits the data?
  We respond this question adopting a visual approach.
  We start considering that
  the shrinkage estimates of the $\delta_i$'s are identical between the two models.
  We thus compute the density plot of the shrinkage estimates (our best estimate of the $\delta_i$'s).
  We take such density as the ground truth (this is actually our best approximation to the ground truth) and we plot it in thick black (Fig.~7).
  Then we sample 8000 $\delta_i$'s from both variants of the model, obtaining two further densities.
  We then plot the three densities for each pair of classifiers (Fig.~7).
  We produce all the density plots using the default kernel density estimation provided in R.
  In general  the hierarchical model, being more flexible, fits better the data than the model equipped with a simple Gamma prior.
  

\subsubsection{Sensitivity on the prior on $\sigma_0$ and $\sigma_i$}\label{sec:sens-sigma}

The model conclusions are  moreover robust with respect to the specification of the priors $p(\sigma_i)$ and $p(\sigma_0)$. Recall that $\sigma_i$ is the standard deviation
on the i-th data set while $\sigma_0$ is the standard deviation of the high-level distribution.

Our model assumes
$\sigma_i \sim \mathrm{unif} (0,\bar{\sigma})$ where 
$\bar{\sigma}= 1000\bar{s}$  where $\bar{s}$ is the average
of the sample standard deviations of the different data sets.
The posterior distributions of the standard deviation obtained with this prior are insensitive
on the choice of $\bar{s}$ as long as $\bar{s}$ is large enough
\citep{kruschke2013bayesian,gelman2006prior}.
Indeed we experimentally verified that we obtained identical posterior estimates of the 
standard deviations adopting as upper bound
$\bar{\sigma}= 1000\bar{s}$ or 
$\bar{\sigma}= 100\bar{s}$.

The same consideration applies to $\sigma_0$, whose prior is 
$p(\sigma_0) =  unif(0,\bar{s_0})$.
We obtain the same posterior distribution for $\sigma_0$
 using  as upper bound
  $\bar{s_0}=1000  s_{\bar{x}}$ or $\bar{s_0}=100  s_{\bar{x}}$, 
  where $s_{\bar{x}}$ is the  standard deviation of the 
  $\bar{x}_i$'s.

\subsection{Comparing the signed-rank and the hierarchical test}\label{sec:uci-signed-rank}
  We compare the conclusions of the hierarchical model and of the signed-rank test
  on the same cases of the previous section.
  The results are given in Tab.~\ref{tab:multiple-dsets-rope}.

\begin{table}[!ht]
 	\rowcolors{3}{white}{lightblue}
 	\centering
 	\begin{tabular}{@{}ccccc}
 		\toprule
 		          & \multicolumn{3}{c}{\textbf{Hierarchical}} & \textbf{Signed-rank} \\
 		pair      & left & rope & right                        & p-value     \\
 		\midrule
 		nbc-hnb   & 1.00 & 0.00 & 0.00                        & 0.00        \\
 		nbc-j48   & 0.80 & 0.02 & 0.18                        & 0.46        \\
 		nbc-j48gr & 0.84 & 0.02 & 0.14                        & 0.39        \\
 		hnb-j48   & 0.03 & 0.10 & 0.87                        & 0.07        \\
 		hnb-j48gr & 0.03 & 0.07 & 0.90                        & 0.08        \\
 		j48-j48gr & 0.00 & 1.00 & 0.00                        & 0.00        \\
 		\bottomrule
 	\end{tabular}
 	\caption{Posterior probabilities of the hierarchical model 
 		and p-values of the signed-rank.
 	\label{tab:multiple-dsets-rope}
 }
 \end{table}  

Both  the signed-rank 
and the hierarchical test claim with 95\% confidence
hnb to be significantly more accurate than nbc.

In the following comparisons apart from the last one,
the two tests do not draw any conclusion with 95\% confidence.
The signed-rank does not reject the null hypothesis, 
while  the hierarchical test does not achieve probability larger than 
95\%. 

When the signed-rank test does not reject the null hypothesis, it draws a non-informative conclusion.
We can instead always interpret the posterior odds yielded by the hierarchical model.
When comparing nbc and j48, there is  a positive evidence 
for right (j48 being more accurate than nbc) 
over left 
and strong evidence 
for right over rope.
We thus conclude that there is positive evidence of j48 being practically more accurate than nbc.
Similarly, we conclude that there is positive evidence of j48gr being practically more accurate than nbc

When comparing hnb and j48, there is strong evidence for right
(hnb being more accurate than j48) over both left  and rope. 
We conclude that there is strong evidence of j48 being practically more accurate than
hnb.
We draw the same conclusion when comparing hnb and j48gr.

    \begin{figure}[!ht]
    	\centering
    	\includegraphics[width=15cm]{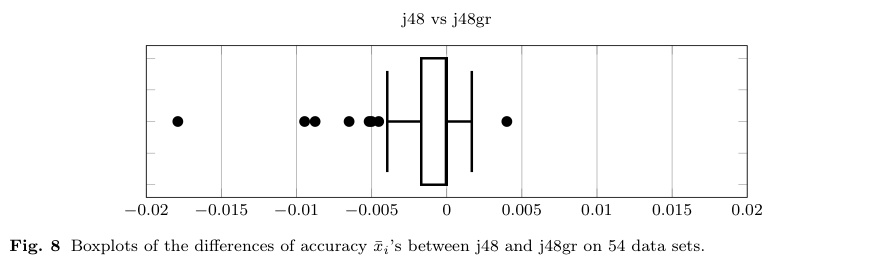}
    \end{figure}
    
The two test draw opposite conclusions when comparing j48 and j48gr.
The signed-rank declares j48gr to be significantly more accurate than j48 (p-value 0.00) while
the hierarchical model declares them to be practically equivalent, with p(rope)=1.
The reason why the two tests achieved opposite conclusions is that the differences have a consistent sign but are small-sized.
In most data sets
the signs of the difference is in favor
of j48gr; this leads 
the signed rank test to claim significance.
Yet the differences lies mostly
within the rope
(Fig.~8). 
The hierarchical model
shrinks them further towards the overall mean and eventually
claims the two classifiers to be practically equivalent. 
The posterior probabilities remain unchanged even adopting the half-sized rope
(-0.005, 0.005). 
Reducing further the rope does not seem meaningful and thus we conclude that, once the magnitude
of the differences is taken into consideration, the accuracy of j48 and j48gr are practically equivalent, even if most signs are in favor of j48gr.

\section{Conclusions}
The proposed approach is  a realistic model of the data generated 
by cross-validation across multiple data sets.
It is the first approach that 
represents both the distribution of the $\delta_i$'s 
across data sets 
and the distribution of the cross-validation results on the i-th data set given $\delta_i$ and $\sigma_i$.

Compared to the signed-rank, it is more conservative as it rejects less easily the null hypothesis.
In  fact the signed-rank assumes a point null hypothesis, while the 
null hypothesis of the hierarchical model contains all the values of the rope.
This is a more realistic null hypothesis which is more difficult to reject.

This allows the hierarchical test 
to detect classifiers which are practically equivalent.
Being more conservative, the hierarchical test is generally less powerful than the signed-rank: when faced with two classifiers which are significantly different, it rejects the null hypothesis with more difficulty.

Yet the interpretation of the posterior odds allows drawing meaningful conclusions even when
the posterior probabilities do not exceed 95\%. 

The hierarchical model yields more accurate estimates of the $\delta_i$'s than the usual maximum likelihood estimator, because it jointly estimates the various $\delta_i$'s applying shrinkage.
Our results show that the shrinkage estimator yields consistently lower mean squared error than the MLE in a variety of challenging experiments.

It is possible to visually verify 
the fit offered by  model by comparing the fitted posterior distribution $p(\delta_i)$ to the density of the shrinkage estimates. In most cases we have found that the Student distribution provides a satisfactory fit.
Yet an interesting research direction is thus the adoption of a non-parametric approach for modelling $p(\delta_i)$,
thus overcoming the Student assumption. This is a non-trivial task which we leave for future research.

      \section*{Acknowledgments}
The research in this paper has been partially supported by the Swiss NSF grants ns.~IZKSZ2\_162188
and n.~200021\_146606.  

  %
  %
  %
  \section{Appendix}
  \subsection{Implementing two classifiers with known difference of accuracy}\label{sec:classifiers-known}

  On the i-th data set
      we need to simulate two classifier whose actual difference of accuracy is 
  $\delta_i$. 
  We start by sampling the instances from a naive Bayes model with two features.
      Let us denote by $C$ the class variables with states are $\{c_0,c_1\}$ and
  by $F$ and $G$ the two features with states $\{f_0,f_1\}$ and $\{g_0,g_1\}$.
  The naive Bayes model is thus $G \leftarrow C \rightarrow F$.
      The parameters of the conditional probability tables are: $P(c_0)$=0.5; $P(f_0|c_0)=\theta_f$; $P(f_0|c_1)=1-\theta_f$;
      $P(g_0|c_0)=\theta_g$; $P(g_0|c_1)=1-\theta_g$ with $\theta_f>0.5$.
      The remaining elements of the conditional probability tables are the complement to 1 of the above elements.
      We set $\theta_f$=0.9 and $\theta_g = \theta_f + \delta_i$.
We sample the data set from this naive Bayes model.

  During cross-validation we train and test  the two competing classifiers $C \rightarrow F$
      and $C \rightarrow G$.
      Their expected accuracies are $\theta_f$ and $\theta_g$ respectively, and thus
      their expected difference of accuracy is 
      $\theta_f-\theta_g = \delta_i$.
      Consider classifier $C \rightarrow F$. 
      Assume that the marginal probabilities of the class have been correctly estimated. The classification thus depends only on the conditional probability of the feature given the class
      If $F$=$f_0$ the most probable class is $c_0$ as long as $\hat{P}(c_0|f_0)>0.5$, where $\hat{P}$ denotes the conditional probability estimated from data.
      The accuracy of this  prediction is $\theta_f$.
      It  $F$=$f_1$, the most probable class is $c_1$ as long as $\hat{P}(c_1|f_1)=\theta_f>0.5$.
      Also the accuracy of this  prediction is $\theta_f$.
      If the bias of conditional probability ($\hat{P}(c_0|f_0)>$0.5 and $\hat{P}(c_1|f_1)>$0.5) is correctly estimated 
      the accuracy of classifier $C \rightarrow F$ on a large test set is $\theta_f$.
      Analogously,  the accuracy of classifier $C \rightarrow G$ in the same conditions is $\theta_g$, so that their difference 
  is $\delta_i$. Since the sampled data set have finite size the  mean difference of accuracy $\bar{x}_i$ measured by cross-validation will fluctuate with some variance around $\delta_i$.
      \subsection{Proofs}
				\paragraph{Proof of Proposition \ref{prop:1}}
  Consider  the  hierarchical model:
  \begin{equation}
  \label{eq:simplfiedmod0}
  \begin{array}{l}
  P(\bm{\bar{x}},\bm{\delta},\delta_0,\sigma_0^2)\\
  =\prod\limits_{i=1}^{q} N(\mathbf{x}_{i};\mathbf{1}\delta_i,\bm{\Sigma}) N(\delta_i;\delta_o,\sigma_o^2)p(\delta_o,\sigma_o^2)\end{array}
  \end{equation}
  We aim at computing the derivative of the $\log(P(\bm{\bar{x}},\bm{\delta},\delta_0,\sigma_0^2))$ w.r.t.\
  the parameter $\delta_i,\delta_0,\sigma_o^2$.
  Consider  the quadratic term from the first and second Gaussian:
  $$
  \frac{1}{2}(\mathbf{x}_{i}-\mathbf{1}\delta_i)^T\bm{\Sigma}^{-1}(\mathbf{x}_{i}-\mathbf{1}\delta_i)+\frac{1}{2\sigma_o^2}(\delta_i-\delta_o)^2;
  $$
  its derivatives w.r.t.\ $\delta_i$ is $  \mathbf{1}^T\bm{\Sigma}^{-1}(\mathbf{x}_{i}-\mathbf{1}\delta_i)+\frac{1}{\sigma_o^2}(\delta_i-\delta_o)$.
  Exploiting the fact that 
  $$
  \begin{array}{l}
  \mathbf{1}^T\bm{\Sigma}^{-1}(\mathbf{x}_{i}-\mathbf{1}\delta_i)=\mathbf{1}^T\bm{\Sigma}^{-1}(\mathbf{x}_{i}-\mathbf{1}\bar{x}_i+\mathbf{1}\bar{x}_i-\mathbf{1}\delta_i)\\
  =\mathbf{1}^T\bm{\Sigma}^{-1}(\mathbf{1}\bar{x}_i-\mathbf{1}\delta_i),
  \end{array}
  $$
  it follows that
  $$
  \frac{d}{\delta_i}\ln(P(\cdot)) \propto \frac{1}{\sigma_n^2}(\bar{x}_i-\delta_i)+\frac{1}{2\sigma_o^2}(\delta_i-\delta_o)^2,
  $$
  where $\sigma^2_n=\tfrac{1}{\mathbf{1}^T\bm{\Sigma}^{-1}\mathbf{1}}=\tfrac{1}{n^2}\mathbf{1}^T\bm{\Sigma}\mathbf{1}$.
  The latter equality can be derived by \cite{coraniML2015}[Appendix], i.e.,
  $$
  \frac{1}{\mathbf{1}^T\bm{\Sigma}^{-1}\mathbf{1}}=\frac{n}{1+(n-1)\rho}=\frac{1}{n^2}\mathbf{1}^T\bm{\Sigma}\mathbf{1}.
  $$
  The other derivatives can be computed easily.
			\paragraph{Proof of Proposition \ref{prop:2}}
  Let us consider the likelihood:
  \begin{equation}
      \begin{array}{l}
      p(\mathbf{x}_i|\delta_i,\bm{\Sigma})=N(\mathbf{x}_i;\mathbf{1}\delta_i,\bm{\Sigma})\\
      =\dfrac{\exp(-\frac{1}{2}(\mathbf{x}_i-\mathbf{1}\delta_i)^{T}\bm{\Sigma}^{-1}(\mathbf{x}_i-\mathbf{1}\delta_i))}{(2\pi)^{n/2}\sqrt{|\bm{\Sigma}|}}.\vspace{2mm}\\
      \end{array}\label{eq:lik-correlated2}
      \end{equation}
  Let us define $\bar{x}_i=\sum_{j=1}^n \mathbf{x}_{ij}/n$.
  The MSE of the maximum likelihood estimator is:
  \begin{align*}
  \nonumber
  \mathrm{MSE_{MLE}}&=\iint \left(\delta_i-\bar{x}_i\right)^2 N(\mathbf{x}_i;\mathbf{1}\delta_i,\bm{\Sigma})p(\delta_i) d\mathbf{x}_i d\delta_i.\\
  \end{align*}
  Consider that  $  \left(\delta_i-\bar{x}_i\right)^2=\left(\delta_i-\tfrac{1}{n}\mathbf{1}^T\mathbf{x}_i\right)^2$
  where $\tfrac{1}{n}\mathbf{1}^T$ is a linear transformation of the variable $\mathbf{x}_i$.
  From the properties of the Normal distribution, it follows that
  $$
  \int \left(\delta_i-\tfrac{1}{n}\mathbf{1}^T\mathbf{x}_i\right)^2 N(\mathbf{x}_i;\mathbf{1}\delta_i,\bm{\Sigma}) d\mathbf{x}_i =\frac{1}{n^2}\mathbf{1}^T\bm{\Sigma}\mathbf{1}
  $$
  and since 
  $$
  \int(\frac{1}{n^2}\mathbf{1}^T\bm{\Sigma}\mathbf{1})p(\delta_i) d\mathbf{x}_i d\delta_i=\frac{1}{n^2}\mathbf{1}^T\bm{\Sigma}\mathbf{1},
  $$
  we derive the first result.
			\paragraph{Proof of Proposition \ref{prop:3}}
  The MSE of the shrunken estimator can be obtained in a similar way.
  First observe that 
  \begin{align}
  \nonumber
  &\left(\delta_i-w\bar{x}_i-(1-w)\delta_0\right)^2\\
  \nonumber
  &=w^2\left(\delta_i-\bar{x}_i \right)^2+(1-w)^2\left(\delta_i-\delta_0 \right)^2 \\
  \nonumber
  &+2w(1-w)\left(\delta_i-\bar{x}_i \right)\left(\delta_i-\delta_0 \right)
  \end{align}
  and its expected value w.r.t.\ $ N(\mathbf{x}_{i};\delta_i,\sigma_n^2)p(\delta_i)$ is:
  \begin{align}
  \nonumber
  &\int \left[w^2\sigma_n^2+(1-w)^2\left(\delta_i-\delta_0 \right)^2 \right]p(\delta_i)  d\delta_i\\
  &=w^2\sigma_n^2+ (1-w)^2\sigma_0^2,
  \end{align}
  where we have denoted  $\sigma_n^2=\frac{1}{n^2}\mathbf{1}^T\bm{\Sigma}\mathbf{1}$.

\clearpage
\bibliography{biblio}
\bibliographystyle{icml2016}

      \end{document}